\documentclass[letterpaper, 10 pt, conference]{ieeeconf}
\IEEEoverridecommandlockouts
\usepackage{gensymb}
\usepackage{textcomp}
\usepackage{graphicx}
\usepackage{amsmath}
\usepackage{cite}
\usepackage{booktabs}  
\usepackage{makecell}   
\usepackage[table]{xcolor}
\usepackage{colortbl}
\usepackage{url}
\usepackage{amssymb}
\usepackage{booktabs}
\usepackage{multirow}
\usepackage{threeparttable}
\usepackage{float}

\title{\LARGE \bf
Musculoskeletal Motion Imitation for Learning Joint-Agnostic Exoskeleton Controllers in Healthy and Impaired Gait
}

\author{Itak Choi$^{*}$, Ilseung Park$^{*}$, Eni Halilaj, and Inseung Kang%
\thanks{This work has been submitted to the IEEE for possible publication. Copyright may be transferred without
notice, after which this version may no longer be accessible.}%
\thanks{$^{*}$These authors contributed equally to this work. (Corresponding author: Itak Choi,
itakc@andrew.cmu.edu)}%
\thanks{All authors are with the Department of Mechanical Engineering, Carnegie Mellon University, Pittsburgh, PA
15213, USA.}%
\thanks{This research was supported by the NIH R21 Award 1R21EB037268-01.}}

\begin{document}

\maketitle
\begin{abstract}
Designing generalizable control policies for lower-limb exoskeletons is often constrained by the need for large biomechanics datasets or lengthy human-in-the-loop optimization.
To address this challenge, we introduce a joint-agnostic framework that combines physiologically plausible musculoskeletal simulation with reinforcement learning to enable scalable exoskeleton assistance for both healthy individuals and those with muscle weakness.
Our base control policy reproduces both the kinematics and kinetics of healthy reference gait.
Fine-tuned from the base policy, the resulting exoskeleton control policies produce assistive torque profiles at the hip, knee, and ankle that align with those validated in human experiments.
For simulated impaired-gait models, the learned control policies yield asymmetric, deficit-specific exoskeleton assistance that improves both metabolic cost and bilateral kinematic symmetry without explicit prescription of the assistance pattern.
We further deployed the learned controller on a robotic hip exoskeleton, where the applied torque closely matched the simulated teacher and reduced the net metabolic rate by 6.9 $\pm$ 6.4\% in three healthy participants.
These results indicate that musculoskeletal motion imitation can potentially serve as a scalable foundation for exoskeleton control across joints and across healthy gait and gait affected by lower-limb muscle weakness, reducing the need for extensive physical trials.
The project page is available at \url{https://ichoi02.github.io/Imitation-Exo/}.
\end{abstract}


\section{Introduction}
Over the past decade, lower-limb wearable robots, such as exoskeletons, have advanced considerably in both their target user populations and objectives \cite{sawicki2020exoskeleton, siviy2023opportunities}.
Initially developed and evaluated for healthy individuals, these systems are now being deployed for people with motor impairments, including those with neurological disorders \cite{gao2025wearable}.
Concurrently, the objectives of these devices have expanded to broader and more clinically relevant outcomes such as correcting crouch gait \cite{lerner2017lower}, improving post-stroke walking \cite{awad2017soft, pruyn2026portable}, averting freezing of gait \cite{kim2024soft}, and personalizing assistance to individual users \cite{ingraham2022role, kang2025online}.

Despite these advances, developing effective exoskeleton controllers remains a central challenge.
Controller design must account for how humans regulate their movements, which is still not fully understood.
To address this challenge, prior studies have employed human-in-the-loop optimization \cite{zhang2017human}, where assistance parameters are iteratively adjusted based on human physiological responses using gradient-free methods.
This approach identifies user- and task-specific assistance profiles that improve human outcomes such as reducing metabolic cost \cite{slade2024human}.
However, because optimization is typically performed for a limited number of users and tasks, the resulting parameters do not readily generalize to diverse locomotor activities.
In addition, the process requires exhaustive search iterations across parameter conditions, making it time-consuming, labor-intensive, and physically demanding, limiting practicality for real-world deployment.

To overcome the task-specific limitations of existing controllers, Molinaro \textit{et al.} proposed a deep-learning-based, task-agnostic paradigm using real-time estimates of physiological states, specifically biological joint moments \cite{molinaro2024task}.
The controller generalizes across diverse locomotor tasks and healthy users, and reduces metabolic cost.
However, training the moment estimator requires large-scale biomechanical experiments, typically conducted in high-fidelity laboratory environments.
Furthermore, while this strategy may improve certain biomechanical metrics, it may not directly address other clinically relevant objectives, such as enhancing stability.

Reinforcement learning using musculoskeletal simulation offers a compelling path to address these limitations.
Training entirely \textit{in silico} enables systematic evaluation across prolonged and diverse locomotor conditions, including pathological gaits that are difficult to study extensively in human experiments.
Building on this potential, recent studies have generated exoskeleton assistance in simulation for models with pathological gait \cite{leem2026exo, yuan2026gait}.
However, their effects were not validated in the real world, and each approach was limited to a single joint.
Conversely, other studies have demonstrated the real-world effect of exoskeleton controllers trained in musculoskeletal simulation \cite{park2026learning, yuan2026smat} on healthy participants.
Yet, these approaches were limited to the hip joint and were not extended to impaired gait.
In addition, several of these methods prescribe the assistance pattern through rewards that favor positive hip power \cite{yuan2026smat, leem2026exo}, rather than letting it emerge from whole-body metabolic cost.
Such assistance may not transfer to joints such as the knee, which primarily exhibits negative power during walking and running \cite{farris2012mechanics}.

\begin{figure*}[t]
    \centering
    \includegraphics[width=1.0\linewidth]{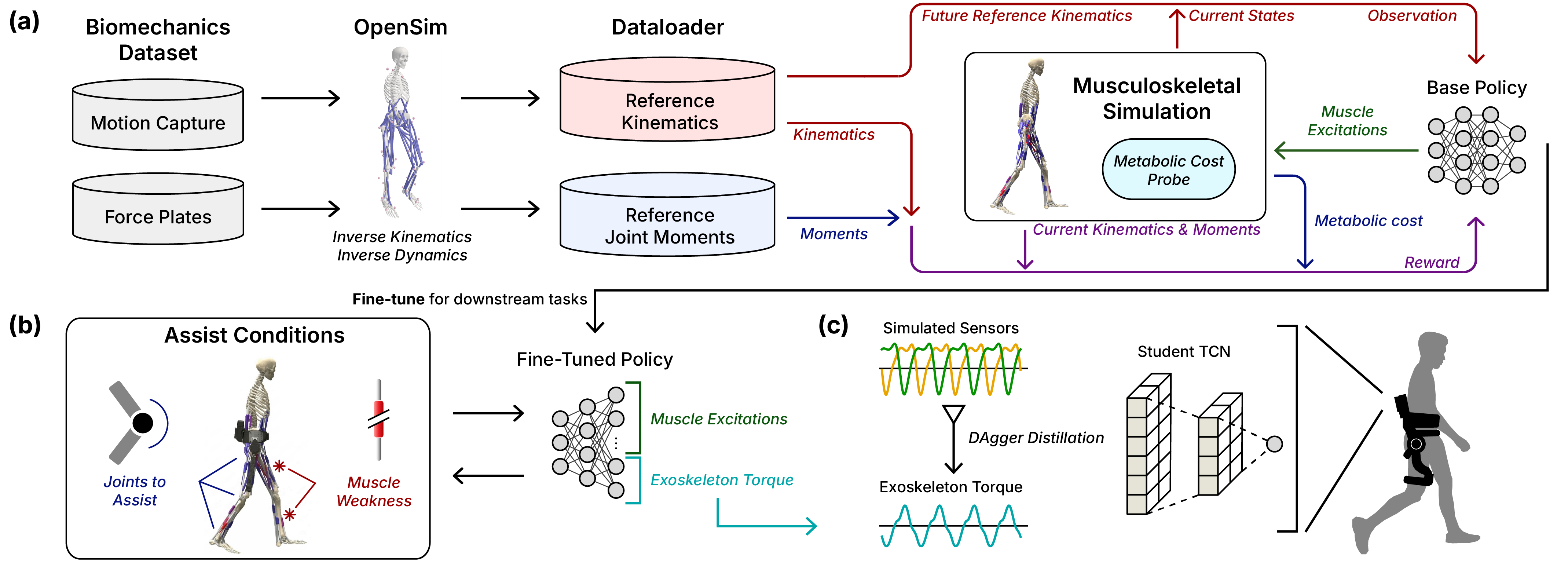}
    \caption{Overview of the musculoskeletal simulation and learning framework.
    (a) Reference kinematics and joint moments are extracted using OpenSim \cite{delp2007opensim} from a biomechanics dataset \cite{scherpereel2023human} containing synchronized motion-capture and force-plate data.
    The policy observes future reference kinematics and outputs muscle excitations to imitate healthy human locomotion.
    The reward function is designed to track reference kinematics and moments while simultaneously minimizing energy expenditure via a metabolic cost probe.
    (b) Fine-tuning phase for the human-exoskeleton system with lower kinematic tracking gains and a higher metabolic penalty.
    The action space is extended to include assistive torque motors alongside muscle excitations to simulate exoskeleton assistance.
    Adjustments to the musculoskeletal model can be applied simultaneously to train impaired-gait-specific assistance.
    (c) Hardware deployment of the fine-tuned teacher policy.
    A student temporal convolutional network (TCN) learns a mapping from onboard sensor readings to exoskeleton torque commands in simulation.
    The trained student network runs on the exoskeleton's onboard computer in real time.}
    \label{fig:method}
\end{figure*}

To our knowledge, no existing framework uses a unified pipeline to train exoskeleton controllers for different joints and muscle weakness.
We show that, in a physiologically plausible simulation, a reward formulation that minimizes metabolic cost while imitating healthy gait leads to emergent assistive torque profiles comparable to those validated in physical experiments.
In summary, the proposed method has three key contributions: (1) muscle-driven motion imitation that reproduces physiologically plausible human kinematics and kinetics; (2) a biomechanics-inspired, joint-agnostic approach for training exoskeleton controllers for both healthy users and individuals with muscle weakness; and (3) simulation and experimental validation of the trained controllers (Fig. \ref{fig:method}).
Together, these provide a step toward scalable exoskeleton assistance, which is particularly valuable for patients who cannot undergo extensive experimental trials.

\section{Methods}
\subsection{Environment}
\subsubsection{Simulation Setup}
The simulation environment was built on the Hyfydy physics engine \cite{geijtenbeek2021hyfydy} with the SCONE Python API \cite{geijtenbeek2019scone}.
Hyfydy is designed for high-fidelity musculoskeletal simulation, executing physics integration while querying the control policy at 25 Hz, with contact forces that include realistic nonlinear damping for both dynamic and viscous friction coefficients.
SCONE provides direct access to the internal states of the model, including kinematics, kinetics, and detailed muscle-level data, which served as inputs to the observation space of the policy.

\subsubsection{Human-Exoskeleton System}
For this study, we used the Hyfydy 3D H2190 model, which features 21 degrees of freedom (DoF) and 90 muscles \cite{schumacher2025emergence}.
To simulate exoskeleton assistance, we augmented the model with ideal torque motors at the bilateral hip, knee, and ankle joints in the sagittal plane.
For conditions with an exoskeleton, device mass was incorporated to enable a more realistic estimation of metabolic cost.
Total mass values were obtained from open-source exoskeleton hardware reported in the literature~\cite{williams2025openexo}, and distributed across the corresponding limb segments using a heuristic approach.
The inertia of each affected segment was then scaled proportionally to reflect the added mass (Table \ref{tab:mass_distribution}).

\begin{table}[htbp]
\centering
\caption{Mass Distribution of Simulated Exoskeleton Devices}
\label{tab:mass_distribution}
\begin{tabular}{l c c c c c}
\toprule

 & \textbf{Pelvis} & \textbf{Thigh} & \textbf{Shank} & \textbf{Foot} & \textbf{Total} \\
 & (kg) & (kg, each) & (kg, each) & (kg, each) & (kg) \\
\midrule
Hip Exo   & 1.5 & 0.7 & --  & --  & 2.9 \\
Knee Exo  & 0.5 & 0.6 & 0.6 & --  & 2.9 \\
Ankle Exo & 1.5 & --  & 1.0 & 0.2 & 3.9 \\
No Exo    & --  & --  & --  & --  & --  \\
\bottomrule
\end{tabular}
\end{table}

\begin{table*}[t]
\centering
\begin{threeparttable}
\caption{Reward and Penalty Definitions, Gain Constants, and Weights}
\label{tab:reward_terms}
\renewcommand{\arraystretch}{1.35}
\begin{tabular*}{\textwidth}{@{\extracolsep{\fill}} l c l c c c c}
\toprule
 & & & \multicolumn{2}{c}{\textbf{Base Policy}} & \multicolumn{2}{c}{\textbf{Fine-Tuning}} \\
\cmidrule(lr){4-5} \cmidrule(lr){6-7}
\textbf{Term} & \textbf{Type} & \textbf{Definition} & $k$ & $w$ & $k$ & $w$ \\
\midrule
Joint position $r^{pos}_t$ & Reward
  & $\exp\{k^{pos}\sum_i(\hat{\theta}^i_t-\theta^i_t)^2\}$
  & $-2.0$ & $0.25$ & $-0.4$ & $0.25$ \\
Joint velocity $r^{vel}_t$ & Reward
  & $\exp\{k^{vel}\sum_i(\hat{\dot{\theta}}^i_t-\dot{\theta}^i_t)^2\}$
  & $-0.05$ & $0.1$ & $-0.01$ & $0.1$ \\
Root trajectory $r^{root}_t$ & Reward
  & $\exp\{k^{root}\|\hat{p}^{root}_t-p^{root}_t\|^2\}$
  & $-500.0$ & $0.15$ & $-500.0$ & $0.15$ \\
End-effector $r^{ee}_t$ & Reward
  & $\exp\{k^{ee}\sum_j\|\hat{p}^j_t-p^j_t\|^2\}$
  & $-80.0$ & $0.25$ & $-16.0$ & $0.25$ \\
Joint torque $r^{torq}_t$ & Reward
  & $\exp\{k^{torq}\sum_i(\hat{\tau}^i_t-\tau^i_t)^2\}$
  & $-2.0$ & $0.25$ & $-0.4$ & $0.25$ \\
Metabolic effort $r^{eff}_t$ & Penalty
  & $E_\text{metabolic}$ 
  & -- & $3\times10^{-5}$ & -- & $3\times10^{-4}$ \\
Smoothness $r^{smt}_t$ & Penalty
  & $\frac{1}{N}\sum_{n=1}^{N}(e^n_t-e^n_{t-1})^2$
  & -- & $1.0$ & -- & $1.0$ \\
Exo energy $r^{exo}_t$ & Penalty
  & $\frac{1}{2}\sum_{s\in\{L,R\}}|\tau^{s,exo}_t|/\tau^{exo}_{max}$
  & -- & -- & -- & $0.1$ \\
\bottomrule
\end{tabular*}
\begin{tablenotes}[flushleft]\footnotesize
\item[] $\hat{x}$: reference value of $x$; 
$i$: rotational DoF; $j$: end-effector (torso and both feet); $n$: muscle ($N=90$); $s$: leg (left, right);
$\theta$: joint angle; $p$: body position; $\tau$: joint moment;
$E_\text{metabolic}$: per-step metabolic rate;
$e$: muscle excitation;
$\tau^{exo}$: exoskeleton torque; $\tau^{exo}_{max}$: exoskeleton torque limit.
\end{tablenotes}
\end{threeparttable}
\end{table*}

\subsection{Reference Motion Processing}
The target reference motion for our framework was obtained from an open-source biomechanics dataset \cite{scherpereel2023human}, which includes synchronized motion-capture and ground reaction force (GRF) data.
For this study, we used treadmill walking data from a single subject as a reference healthy gait, consisting of an approximately two-minute clip covering five speeds ranging from 0.6 to 2.2 m/s during walking and running.
The treadmill data were first converted to an overground reference frame by adding the treadmill belt velocity.
Reference joint angles and moments were then computed using inverse kinematics and inverse dynamics in OpenSim \cite{delp2007opensim}, along with positions of key body segments such as the head, pelvis, and feet.
To encourage bilateral symmetry in the learned policy, the clip was duplicated and mirrored across the sagittal plane, effectively augmenting the training dataset.

\subsection{Deep Reinforcement Learning}
The training framework is formulated as a Markov decision process $\mathcal{M} = \langle S, A, P, r, \gamma \rangle$, in which a stochastic neural-network policy $\pi(a_t|s_t)$ samples an action $a_t \in A$ given the current state $s_t \in S$.
The Hyfydy physics engine then integrates the system dynamics to the next control step, where the environment computes the scalar reward $r_t$ and updates the state to $s_{t+1}$.

\subsubsection{Action Space}
The action space of the human-exoskeleton system is defined as $A = [-1, 1]^{96}$, consisting of muscle excitations $A^m$ and exoskeleton torques $A^e$.
The muscle action space $A^m = [-1, 1]^{90}$ is linearly mapped to $[0, 1]^{90}$, directly representing the excitation levels of all 90 muscles.
Within the Hyfydy integrator, these excitations are converted to muscle activations through first-order forward Euler integration, capturing the physiological delay inherent in muscle activation dynamics.
The exoskeleton action space $A^e = [-1, 1]^{6}$ represents the normalized torques applied to the bilateral hip, knee, and ankle joints in the sagittal plane.
Each exoskeleton can be activated through a configuration setting for a fine-tuning run to generate assistive torque (Fig. \ref{fig:method}b) and is otherwise clamped to zero, as in base policy training (Fig. \ref{fig:method}a).

\subsubsection{Observation Space}
The observation space $S=\mathbb{R}^{\text{839}}$ consists of the current model state $S^c$ and kinematic deviations relative to future reference motion $S^f$.
The current state $S^c$ includes physiological muscle data (fiber length, velocity, force, and activation) alongside GRFs, joint angles, joint moments, linear velocities of the pelvis, and relative positions of key bodies (head and feet) to the pelvis.
The future reference component $S^f$ encodes the sequence of joint angle changes over the next five reference frames, expressed as $S^f = \{\Theta_{t+k} - \Theta_t\}_{k=1}^{5}$, where $\Theta$ denotes a flattened vector of joint angles, angular velocities, and key body positions and velocities.
The complete observation space is thus defined as $S=\{S^c, S^f\}$.

\subsubsection{Reward}
The reward function combines reward terms that encourage kinematic and kinetic tracking of the reference motion with penalty terms that discourage high metabolic cost, abrupt changes in muscle excitation, and excessive exoskeleton torque.
At each time step $t$, the scalar reward $r_t$ is computed as the weighted sum of five reward terms minus three penalty terms, as defined in \eqref{reward}.

\begin{equation}
\label{reward}
\begin{split}
r_t=w^{pos}r^{pos}_{t}+w^{vel}r^{vel}_{t}+w^{root}r^{root}_{t}+w^{ee}r^{ee}_{t}\\
+w^{torq}r^{torq}_{t}-w^{eff}r^{eff}_{t}-w^{smt}r^{smt}_{t}-w^{exo}r^{exo}_{t}
\end{split}
\end{equation}

Table \ref{tab:reward_terms} lists the type, definition, gain constant $k$, and weight $w$ of each term.
The first five terms reward tracking of the reference motion, inspired by DeepMimic \cite{peng2018deepmimic}.
Each is an exponential of the squared error between the reference and the simulated agent, so that it is bounded within $(0, 1]$ and its gain $k$ sets the tolerance to tracking error.
We added the joint torque term $r^{torq}_t$ to the original DeepMimic tracking reward, which compares the muscle-generated joint moments of the agent with the joint moments obtained from inverse dynamics of the reference motion.
The remaining three terms are penalties.
Because minimizing metabolic cost is considered a fundamental principle of human locomotion, the metabolic effort term $r^{eff}_t$ penalizes whole-body gross metabolic power, computed with a metabolic probe based on the model of Umberger \textit{et al.} \cite{umberger2003model, umberger2010stance}.
The smoothness term $r^{smt}_t$ penalizes changes in muscle excitation between consecutive control steps to encourage realistic activation patterns.
The exoskeleton energy term $r^{exo}_t$ is active only during fine-tuning and penalizes the magnitude of the assistive torque.
All gain constants and weights were determined empirically.

\subsection{Policy Training}
The training process consists of two distinct phases.
First, a base policy is trained to reproduce human-like motion in both joint kinematics and kinetics.
Policies are fine-tuned by warm-starting from the base policy to produce assistive torque at the required joints, including under altered conditions such as muscle weakness.

\subsubsection{Healthy Gait}
All policies were trained (Fig. \ref{fig:method}a) using Soft Actor-Critic (SAC) \cite{haarnoja2018soft}, as implemented in the Stable-Baselines3 library \cite{raffin2021stable}.
The base policy was trained across 192 parallel environments for a total of 600 million steps on a desktop workstation equipped with an NVIDIA RTX 5090 GPU and an Intel Ultra 9 285K CPU, requiring approximately 29 hours (Table \ref{tab:sac_parameters}).

\begin{table}[htbp]
\centering
\caption{SAC Hyperparameters}
\label{tab:sac_parameters}
\begin{tabular}{l c}
\toprule
\textbf{Parameter} & \textbf{Value} \\
\midrule
Hidden Layers (Actor / Critic) & [512, 512, 256] \\
Batch Size & 256 \\
Learning Rate & Linear decay from $3 \times 10^{-4}$ \\
Soft Update Coefficient & 0.02 \\
Entropy Coefficient & auto \\
Discount Factor & 0.95 \\
Train Frequency & 4 \\
Gradient Steps & 4 \\
Target Update Interval & 1 \\
Total Steps (Base Policy) & 600 million \\
Total Steps (Fine-Tune) & 150 million \\
\bottomrule
\end{tabular}
\end{table}

The policy and Q-networks are multi-layer perceptrons with three hidden layers of 512, 512, and 256 units.
Each training episode begins by uniformly sampling a starting time from the motion clip.
The model's joint positions and velocities are then initialized to match the corresponding reference state.
At initialization, the vertical position of the model is adjusted such that the net GRF equals the total body weight.
At each control step (25 Hz), the policy processes the current observation and samples excitation values for all 90 muscles.
The simulation then advances to the subsequent control time step, at which point the reward and next observation are computed.
Episodes are truncated to a maximum of 250 steps, corresponding to 10 seconds of simulation time.
An early termination condition is triggered if the model's root position deviates by more than 0.8 m in Euclidean distance from the reference root position.
For tracking robustness, Gaussian offsets are progressively added to the reference joint angles throughout the base policy training, with a maximum offset of 0.07 rad.

\subsubsection{Exoskeleton Control}
Starting from the trained base policy weights, an additional 150 million training steps were conducted for each device condition, activating the extended action space for exoskeleton torque (Fig. \ref{fig:method}b).
To account for the kinematic changes induced by assistive torque, the gain constants for several tracking terms ($k^{pos}, k^{vel}, k^{ee}, k^{torq}$) were reduced and the metabolic penalty ($w^{eff}$) was increased during fine-tuning (Table \ref{tab:reward_terms}).
The mass and inertia of the devices were incorporated into the corresponding limb segments to reduce the sim-to-real gap.

The policy output for exoskeleton torque, bounded within $[-1, 1]$, is first linearly mapped to $[-\tau^{exo}_{max}, \tau^{exo}_{max}]$.
The output then passes through a first-order low-pass filter with a cutoff frequency of 2 Hz.
We set $\tau^{exo}_{max}$ to 30 Nm for the hip and knee, and 60 Nm for the ankle.
The resulting torque is applied directly to the respective joint to assist the muscle-driven model.

To obtain smooth bilateral assistance, we added two terms to the actor loss only on the exoskeleton action dimensions, as defined in \eqref{actor_loss}.
These terms, defined in \eqref{exo_losses}, penalize abrupt changes in assistive torque \cite{mysore2021regularizing} and asymmetric assistance, respectively.

\begin{equation}
\label{actor_loss}
\mathcal{L}_\pi=\mathcal{L}_{SAC}+\lambda_{smt}\mathcal{L}^{exo}_{smt}+\lambda_{sym}\mathcal{L}^{exo}_{sym}
\end{equation}

We used $\mu(s)$, the deterministic action that runs at deployment, to avoid exploration noise increasing $\mathcal{L}^{exo}_{smt}$.
$M_a$ and $M_s$ are operations that mirror actions and observations across the sagittal plane, respectively.
We used $\lambda_{smt}=0.05$ and $\lambda_{sym}=0.5$ for healthy models, and $\lambda_{sym}=0$ for impaired models.

\begin{equation}
\label{exo_losses}
\begin{split}
& \mathcal{L}^{exo}_{smt} = \mathbb{E}_s||\mu(s_t)-\mu(s_{t-1})||^2 \\
& \mathcal{L}^{exo}_{sym} = \mathbb{E}_s||\mu(s)-M_a(\mu(M_s(s)))||^2
\end{split}
\end{equation}

\begin{figure*}[t]
    \centering
    \includegraphics[width=1.0\linewidth]{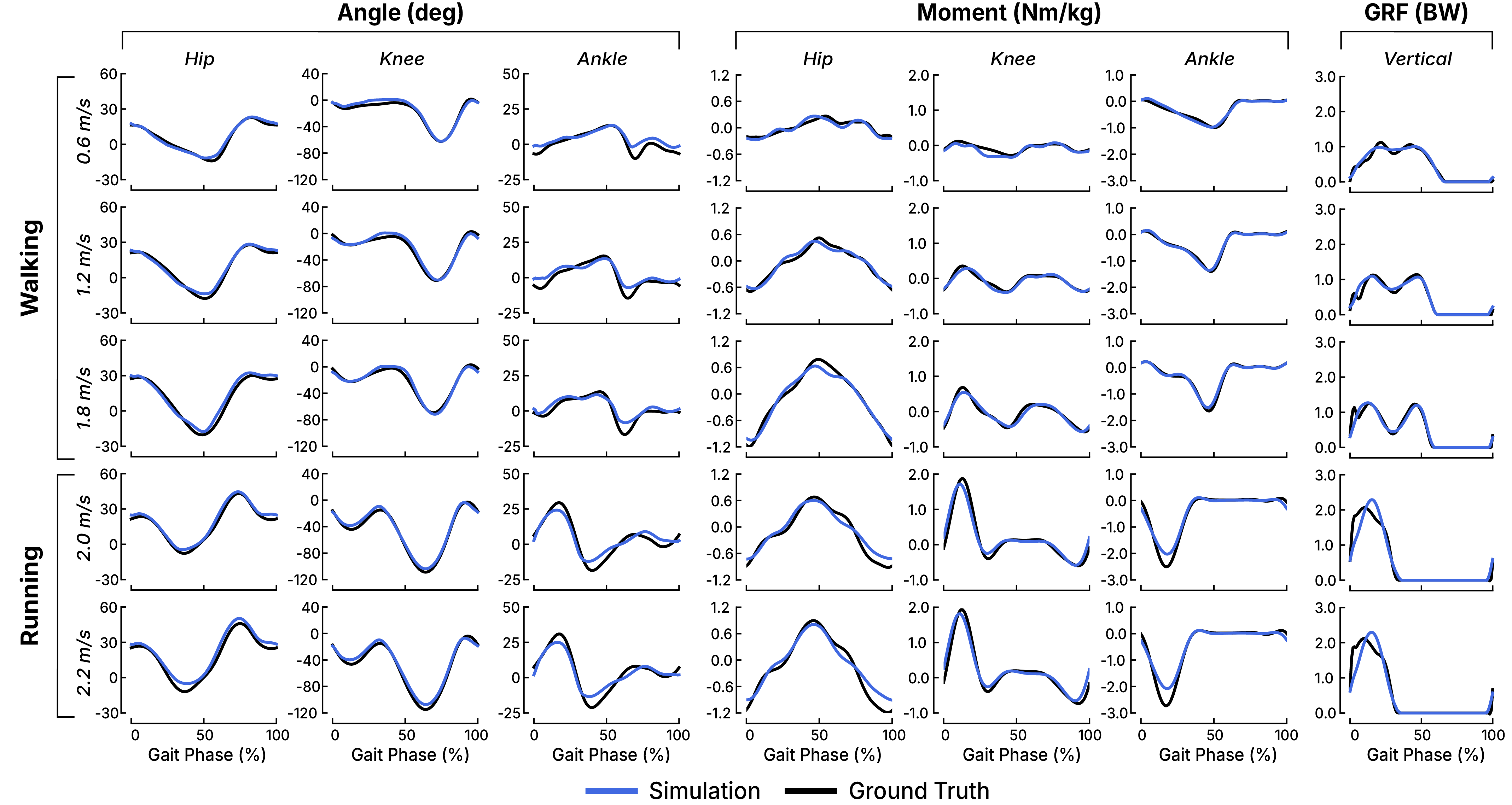}
    \caption{Kinematic and kinetic tracking performance of the base policy across five locomotor speeds (walking and running).
    Blue and black lines represent the simulated agent and the ground-truth data, respectively.
    Joint angles and moments for the hip, knee, and ankle are compared against the ground truth; ground reaction force is not included as a reference signal during training.
    All values are segmented into gait cycles using heel contact events, interpolated to 100 data points, and averaged over 10 cycles.
    A fourth-order zero-lag Butterworth low-pass filter with a 6 Hz cutoff frequency is applied to the ground-truth joint moments to reduce noise.}
    \label{fig:tracking}
\end{figure*}

\subsubsection{Exoskeleton Control for Impaired Gait}
We also trained controllers for impaired gait, with targeted muscle groups artificially weakened.
By simultaneously weakening these muscles and activating the corresponding exoskeleton action space, our framework can generate assistive torque commands tailored for individuals exhibiting such muscular deficits.
In this work, we simulated unilateral plantarflexor and hip flexor weakness.
We chose these two deficits because they are both common and functionally disabling in impaired gait \cite{jonkers2009capacity}, and paired each with the exoskeleton that most directly restores the affected function.
These conditions were simulated by reducing the maximum excitation of every muscle in the affected group to 5\% of its original value.
The aim is not to synthesize pathological gait but to generate appropriate assistance.
We hypothesized that training the weakened model to imitate healthy gait with exoskeleton assistance produces torque profiles that effectively compensate for the underlying impaired gait mechanics.

\subsection{Hardware Deployment}
For hardware validation, we developed a 4.0 kg robotic hip exoskeleton that applied bilateral torques up to 18 Nm.
Each hip was driven by a quasi-direct-drive brushless DC motor (AK80-9, CubeMars, China), and inertial measurement units (IMUs; ICM-20948, Adafruit Industries, USA) on the distal thigh bars measured angular velocities and accelerations.
A 22.2 V, 3300 mAh lithium-polymer battery (HRB, China) powered the system, and an embedded computer (Jetson Orin Nano, NVIDIA, USA) ran the controller onboard (Fig. \ref{fig:experiments}a).

To deploy the exoskeleton controllers, we distilled the hip torque commands of the trained policies into a student network that uses only the sagittal angular velocity of both thighs, similar to the method proposed by Park \textit{et al.} \cite{park2026learning} (Fig. \ref{fig:method}c).
The student $g$ is a temporal convolutional network (TCN) that maps a 24-frame history of [right, left] thigh angular velocity to the torque command for the right hip.
At deployment, two copies of this network serve the two legs: $\tau_R=g(\omega_R,\omega_L)$ and $\tau_L=g(\omega_L,\omega_R)$.
The network has three residual blocks, each with 32 channels, a kernel size of 3, and two weight-normalized causal convolutions, with dilations of 1, 2, and 4.
We trained $g$ with DAgger \cite{ross2011reduction} for eight iterations.
At each control step of iteration $i$, the student commanded the hip exoskeleton with probability $1-0.6^i$, while the teacher always drove the muscles and supplied the correct label.
After each iteration, the student was retrained from scratch on all data collected so far, with AdamW (learning rate $10^{-3}$, weight decay $10^{-3}$, batch size 256, at most 300 epochs).
To make the student robust at deployment, we added Gaussian gyroscope noise and a per-episode bias during episode rollouts, and 2.5\% stationary samples with zero angular velocity mapped to zero torque.

On the hardware, the student output passed through the same 2 Hz filter as in simulation.
We increased the resulting torque by a gain of 1.5 to account for torque transmission loss between the exoskeleton and the human hip joint.

\subsection{Human-Subject Experiments}
Three healthy adults (AB01--AB03; all male, age 26.7 $\pm$ 3.0 years, height 1.76 $\pm$ 0.03 m, and mass 78.7 $\pm$ 6.0 kg; reported as mean $\pm$ SD throughout) participated in the experiment.
They gave written informed consent to a protocol approved by the Institutional Review Board of Carnegie Mellon University (STUDY2024\_00000506).
Each participant first walked wearing the exoskeleton for up to 5 min for familiarization.
We then collected exoskeleton torque profiles across varying speeds.
The speed of an instrumented treadmill (FIT5, Bertec, USA) increased from 0.3 to 2.4 m/s in 0.3 m/s increments, each lasting 15 s.
The commanded torque was normalized by body mass and time-normalized to the gait cycle.
Next, we measured metabolic cost in two conditions, both with the device worn: without assistance (zero torque applied) and with assistance.
After a standing baseline (3 min), participants walked on the treadmill at 1.2 m/s in four 6-min bouts in an ABBA order.
Participants rested 4 min before each bout.
Breath-by-breath measurements (K5, COSMED, Italy) of $\dot V_{O_2}$ and $\dot V_{CO_2}$ were converted to metabolic power \cite{brockway1987derivation}.
The power was averaged over the last 3 min of each bout and normalized by body mass, and the standing rate was subtracted to obtain the net metabolic rate.
Each participant's two bouts per condition were averaged.

\section{Results}
\subsection{Physiological Plausibility}
The simulated agent achieved high-fidelity kinematic and kinetic tracking performance relative to the ground-truth data (Fig. \ref{fig:tracking}).
Evaluated on each gait cycle (10 cycles at each of the five speeds), the model tracked hip, knee, and ankle angles with root mean square errors (RMSE) of 3.15 $\pm$ 1.04$\degree$, 4.42 $\pm$ 1.17$\degree$, and 3.97 $\pm$ 0.77$\degree$, respectively, and corresponding coefficients of determination ($R^2$) of 0.96 $\pm$ 0.02, 0.97 $\pm$ 0.01, and 0.82 $\pm$ 0.09.
For joint kinetics, moment tracking yielded RMSE values of 0.13 $\pm$ 0.05, 0.13 $\pm$ 0.05, and 0.15 $\pm$ 0.09 Nm/kg for the hip, knee, and ankle, respectively, with corresponding $R^2$ values of 0.89 $\pm$ 0.07, 0.78 $\pm$ 0.28, and 0.94 $\pm$ 0.04.
The model also reproduced GRF with an RMSE of 0.15 $\pm$ 0.07 body weights and an $R^2$ of 0.93 $\pm$ 0.03, despite the agent having no access to GRF data during training.

\subsection{Exoskeleton Control for Healthy Gait}

\begin{figure}[t]
    \centering
    \includegraphics[width=1.0\linewidth]{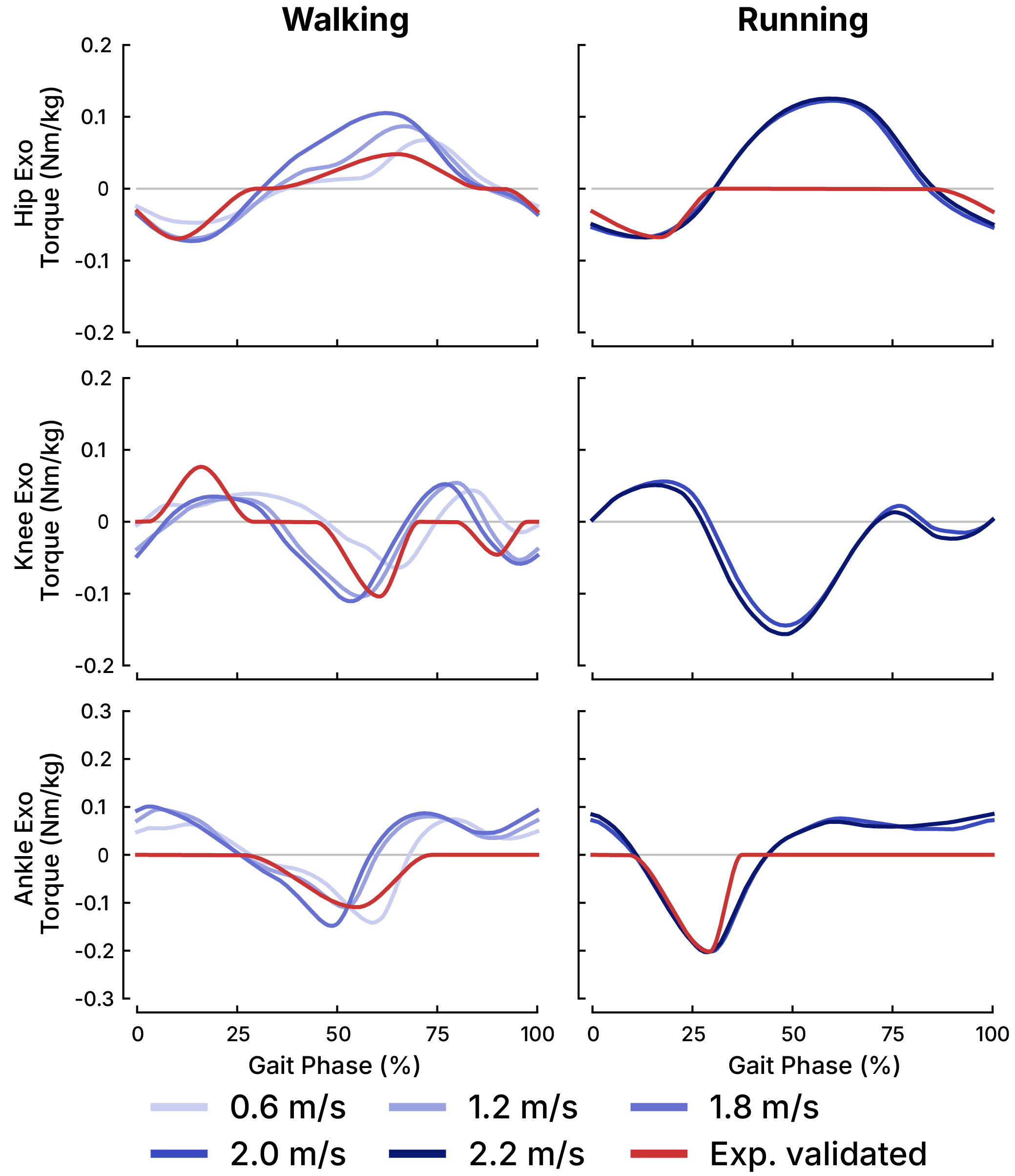}
    \caption{Generated exoskeleton assistive torque profiles (Nm/kg, normalized by model mass) for the hip, knee, and ankle joints during walking and running.
    The assistance profiles from the fine-tuned policies (blue gradients, ranging from 0.6 to 2.2 m/s) are compared against experimentally validated torque profiles from prior work (red) \cite{witte2020improving, bryan2021optimized, kim2019reducing}.
    Reference profiles for running with a hip exoskeleton consist of optimized extension-only splines, while those for the ankle consist of plantarflexion-only splines.
    No optimized knee exoskeleton profile for running has been reported.
    For comparison of peak timings, the maximum amplitude of the optimized reference profiles is normalized to match the negative peak torque of the simulated profiles at the closest corresponding speed.}
    \label{fig:healthy_exo}
\end{figure}

\begin{figure}[t]
    \centering
    \includegraphics[width=1.0\linewidth]{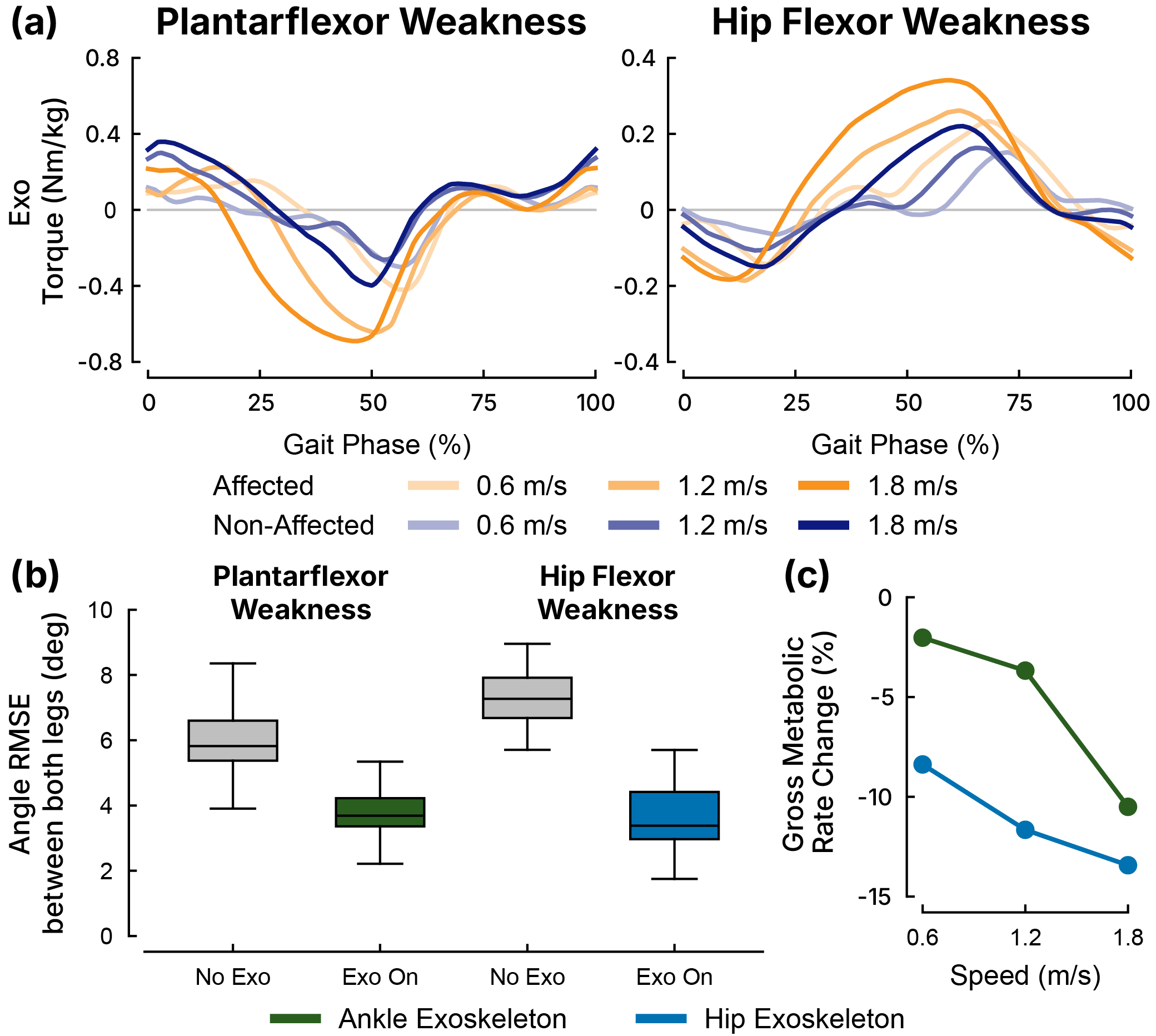}
    \caption{Assistance trained for different muscle weakness and their effect on kinematics and metabolic cost.
    (a) Exoskeleton torque profiles for the affected (orange) and non-affected (blue) sides of the impaired-gait models at walking speeds of 0.6, 1.2, and 1.8 m/s (lighter to darker), each averaged over 10 gait cycles.
    (b) RMSE between the affected and non-affected joint angles without (gray) and with (green and blue) exoskeleton assistance, computed on each of 10 gait cycles at each speed and pooled across speeds.
    (c) Change in gross metabolic cost with exoskeleton assistance relative to the no-exoskeleton baseline averaged at each walking speed, for plantarflexor weakness (green) and hip flexor weakness (blue).}
    \label{fig:impaired_exo}
\end{figure}

The torque profiles from policy rollouts were compared against profiles that reduced metabolic cost in human experiments (Fig. \ref{fig:healthy_exo}) \cite{bryan2021optimized, witte2020improving, kim2019reducing}.
Compared to the human-in-the-loop optimized profiles at 1.25 m/s by Bryan \textit{et al.} \cite{bryan2021optimized}, the simulated hip, knee, and ankle profiles at 1.2 m/s had Pearson's $r$ of 0.95, 0.67, and 0.80, respectively.
To isolate the effect of assistance, the baseline without an exoskeleton was fine-tuned from the same base policy with the same reward and number of training steps, but without the added device mass or assistive torque.
Compared to the model without the exoskeleton, the simulated gross metabolic rate changed by $-$1.5\% with the hip exoskeleton, $-$1.9\% with the knee exoskeleton, and +0.7\% with the ankle exoskeleton during walking (mean of 0.6, 1.2, and 1.8 m/s).
During running (mean of 2.0 and 2.2 m/s), it changed by $-$1.7\% with the hip exoskeleton, +1.2\% with the knee exoskeleton, and 0.0\% with the ankle exoskeleton.
The metabolic rates were computed with the added mass of each exoskeleton device, excluding the constant basal rate.

\subsection{Exoskeleton Control for Impaired Gait}
The learned policy generated asymmetric exoskeleton assistance that targeted the affected side (Fig. \ref{fig:impaired_exo}a).
Computed on each of 10 gait cycles at each speed and pooled across all speeds, the RMSE between affected and non-affected joint angles decreased from 6.01 $\pm$ 1.32$\degree$ to 3.71 $\pm$ 0.76$\degree$ for the plantarflexor weakness model and from 7.38 $\pm$ 1.01$\degree$ to 3.60 $\pm$ 0.95$\degree$ for the hip flexor weakness model (Fig. \ref{fig:impaired_exo}b).
The exoskeleton assistance also reduced gross metabolic cost, averaged at each speed, by 2.0\%, 3.7\%, and 10.5\% for the plantarflexor weakness model and by 8.4\%, 11.7\%, and 13.4\% for the hip flexor weakness model at 0.6, 1.2, and 1.8 m/s, respectively (Fig. \ref{fig:impaired_exo}c).
None of the impaired models, with or without exoskeleton assistance, completed running motion from 2.0 m/s.
The impaired-gait results are therefore limited to walking speeds.

\subsection{Hardware Experiment Outcomes}
\begin{figure}[t]
    \centering
    \includegraphics[width=1.0\linewidth]{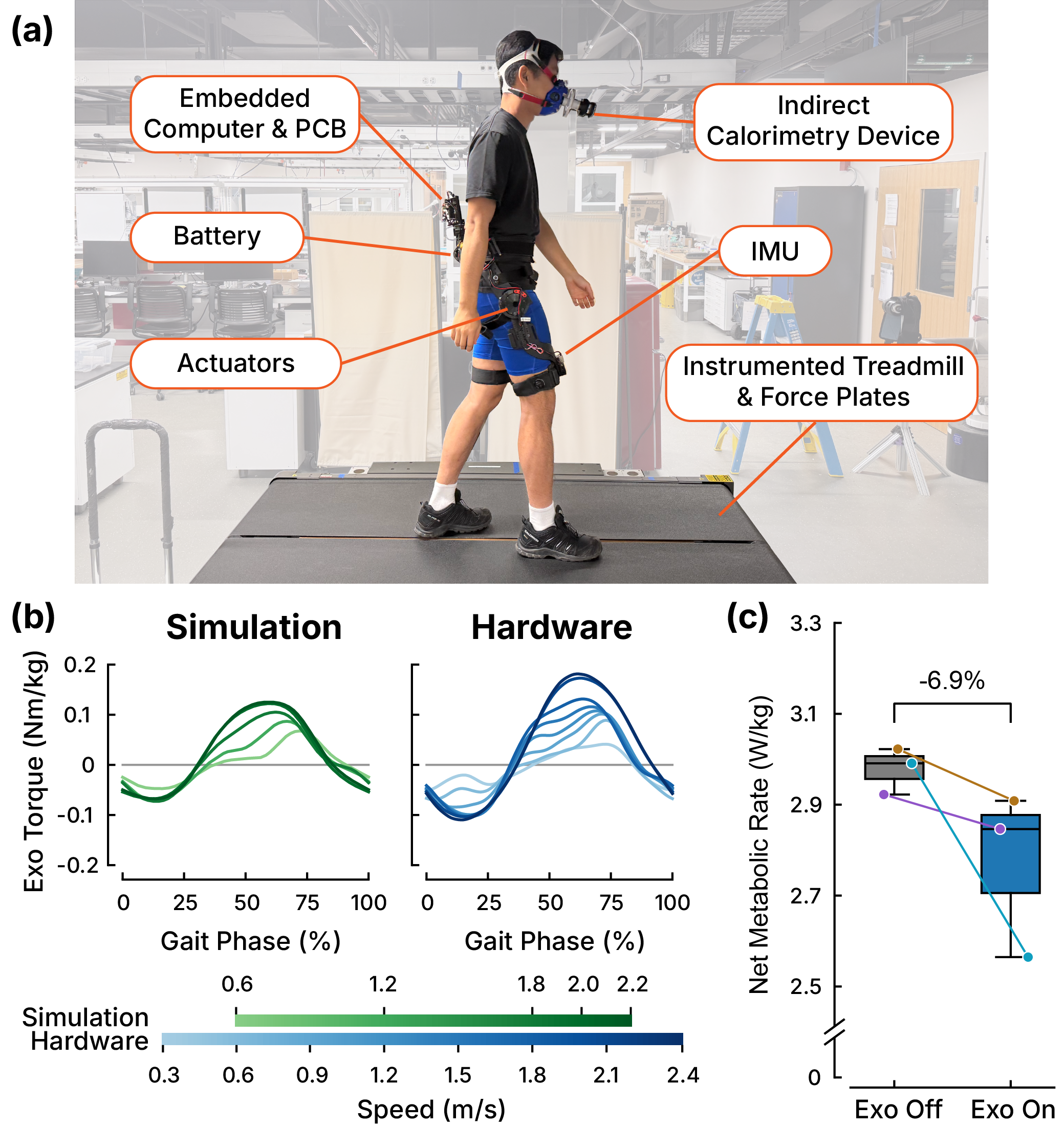}
    \caption{Hardware validation of a trained controller.
    (a) The bilateral hip exoskeleton assists the user.
    The participants were asked to walk on an instrumented treadmill with embedded force plates while wearing an indirect calorimetry device.
    (b) Torque profiles of the teacher policy from simulation (left) and the applied torque from the student TCN on the hardware (right). (c) Comparison of the net metabolic rate with assistance on and off, both with the device worn.}
    \label{fig:experiments}
\end{figure}

At treadmill speeds of 0.6, 1.2, and 1.8 m/s, the applied motor torque, divided by the gain of 1.5, matched the teacher torque in simulation, with RMSEs of 0.013 $\pm$ 0.004, 0.011 $\pm$ 0.001, and 0.013 $\pm$ 0.002 Nm/kg and $R^2$ values of 0.85 $\pm$ 0.09, 0.95 $\pm$ 0.01, and 0.95 $\pm$ 0.02, respectively (n = 3).
At unseen speeds, the student's profiles lay between those of the adjacent speeds (Fig. \ref{fig:experiments}b).
This held both within the trained speed range and outside it, showing that the student interpolated and extrapolated across speeds.
At a treadmill speed of 1.2 m/s, assistance reduced the net metabolic rate by 6.9 $\pm$ 6.4\% compared to walking without assistance (n = 3; Fig. \ref{fig:experiments}c).
The metabolic rate decreased in all three participants, but no statistical analysis was performed due to the small sample size.

\section{Discussion}
While previous studies have explored various approaches to generating optimal exoskeleton assistance \cite{zhang2017human, molinaro2024task}, this framework is distinct in that assistive torque profiles can be generated entirely within simulation for different targeted joints and muscular deficits.
Assistance learned by imitating healthy gait while minimizing metabolic cost also reduced metabolic cost on a physical hip exoskeleton.
For the impaired models, the learned assistance improved kinematic symmetry and reduced metabolic cost in simulation.

Using same reward formulations, the policy generated hip, knee, and ankle assistive torque profiles comparable to those reported in prior experimental studies \cite{bryan2021optimized, witte2020improving, kim2019reducing}, without iterative optimization involving human participants.
Furthermore, we demonstrated sim-to-real transfer of the trained controller on a robotic hip exoskeleton, which reduced metabolic cost relative to walking without assistance.
This reduction was smaller than those reported in previous studies \cite{kim2022reducing, molinaro2024task, yuan2026smat}, and no statistical analysis was performed, as the objective of the experiment was to demonstrate transfer to hardware rather than to maximize the benefit.

Furthermore, we hypothesized that training a weakened model to imitate healthy gait with exoskeleton assistance would produce torque profiles that compensate for the underlying deficit.
The simulation results support this hypothesis: the learned asymmetric assistance reduced metabolic cost and improved bilateral kinematic symmetry relative to walking without the exoskeleton.
Although these results are limited to simulation, they suggest that the learned policies can improve both kinematic symmetry and metabolic cost under targeted muscular deficits.

However, several limitations should be acknowledged.
Although our framework generated control policies across different configurations, only hip assistance was validated on physical hardware.
This hardware experiment also included only three healthy participants without statistical analysis, and it compared with the assistance turned off, not without the device.
In addition, the improved gait of the impaired models was demonstrated only in simulation.
Furthermore, the simulated exoskeletons were modeled as ideal actuators with full torque transmission.
On hardware, the commanded torque was therefore increased by a fixed gain of 1.5 to compensate for transmission losses between the device and the hip joint.
Future work should deploy the controller on knee and ankle exoskeletons and incorporate actuator dynamics and human--device interface compliance into the simulation.
Finally, although the simulation results indicate potential benefits for impaired populations, experiments involving patients are ultimately required to establish clinical validity in real-world settings.

\section{Conclusion}
This study presented a joint-agnostic exoskeleton control framework trained through physiologically plausible musculoskeletal simulation and adaptable to impaired gait.
The base policy accurately tracked the kinematics and kinetics of healthy reference gait from a single subject.
After fine-tuning with assistance enabled, the policy generated hip, knee, and ankle torque profiles whose trends matched those validated in human experiments.
The trained hip controller transferred to a physical exoskeleton and reduced metabolic cost relative to walking without assistance.
For the impaired models, the policy generated asymmetric assistance that improved kinematic symmetry and reduced metabolic cost in simulation.
Because the same reward formulation applies across joints and musculoskeletal conditions, the framework has the potential to scale to other locomotor tasks and user populations, although this was not tested here.
While this study focused on level-ground walking and running, future work should extend the framework to other locomotor tasks and validate the trained controllers in clinical populations.

\bibliographystyle{ieeetr}
\bibliography{references.bib}

\end{document}